%% file: workshop_arxiv.tex
\documentclass[final]{cvpr}

\input{math_commands.tex}

\usepackage{url}

\usepackage{graphicx}
\usepackage{multirow}
\usepackage{subcaption}
\usepackage{dblfloatfix}
\usepackage{booktabs}
\usepackage{colortbl}
\usepackage{xcolor}
\definecolor{mygray}{gray}{.9}
\definecolor{mygreen}{RGB}{34,139,34}

\newcommand{\e}[1]{{\color{black}{\@#1}}}
\newcommand{\zt}[1]{{\color{black}{\@#1}}}

\newcommand{\first}[1]{{\color{blue}{\@#1}}}
\newcommand{\second}[1]{{\color{mygreen}{\@#1}}}

\newcommand{\secvspace}{\vspace{-0.0em}}
\newcommand{\subsecvspace}{\vspace{-0.0em}}
\newcommand{\figvspace}{\vspace{-0.2em}}
\newcommand{\captionvspace}{\vspace{-0.1em}}

\usepackage{times}
\usepackage{epsfig}
\usepackage{graphicx}
\usepackage{amsmath}
\usepackage{amssymb}
\usepackage{cite}

\usepackage[pagebackref=true,breaklinks=true,colorlinks,citecolor=mygreen,bookmarks=false]{hyperref}
\usepackage{cleveref}
\crefname{section}{§}{§§}
\Crefname{section}{§}{§§}

\def\cvprPaperID{35} 
\def\confYear{CVPR 2021}

\begin{document}

\title{Shot in the Dark: Few-Shot Learning with No Base-Class Labels}

\author{Zitian Chen\qquad Subhransu Maji\qquad Erik Learned-Miller\\
University of Massachusetts Amherst\\  {\tt\small \{zitianchen,smaji,elm\}@cs.umass.edu}}

\maketitle

\pagestyle{empty}  
\thispagestyle{empty} 
\begin{abstract}

Few-shot learning aims to build classifiers for new classes from a small number of labeled examples and is commonly facilitated by access to examples from a distinct set of ‘base classes’. The difference in data distribution between the test set (novel classes) and the base classes used to learn an inductive bias often results in poor generalization on the novel classes. To alleviate problems caused by the distribution shift, previous research has explored the use of unlabeled examples from the novel classes, in addition to labeled examples of the base classes, which is known as the \textbf{transductive setting}. In this work, we show that, surprisingly, off-the-shelf self-supervised learning outperforms transductive few-shot methods by 3.9\% for 5-shot accuracy on \emph{mini}ImageNet \textbf{without using any base class labels}. This motivates us to examine more carefully the role of features learned through self-supervision in few-shot learning. Comprehensive experiments are conducted to compare the transferability, robustness, efficiency, and the complementarity of supervised and self-supervised features.
  
\end{abstract}


\secvspace
\section{Introduction}
\secvspace

Deep architectures have achieved significant success in various vision tasks including image classification and object detection. Such success have relied heavily on massive numbers of annotated examples. However, in real-world scenarios, we are frequently unable to collect enough labeled examples. This has motivated the study of few-shot learning (FSL), which focuses on building classifiers for novel categories from one or very few labeled examples. 

Previous approaches to FSL include meta-learning and metric learning. Meta-learning aims to learn task-agnostic knowledge that improves optimization. Metric learning focuses on learning representations on base categories that can generalize to novel categories. Most previous FSL methods attempt to borrow a strong inductive bias from the \textit{supervised} learning of base classes. However, the challenge of FSL is that a {\em helpful} inductive bias, i.e., one that improves performance on novel classes, is hard to develop when there is a large difference between the base and novel classes. 

To address this challenge, previous research explores using unlabeled examples from novel classes to improve the generalization on novel classes, which is referred to transductive few-shot learning. Typical transductive few-shot learning (TFSL) methods include exploiting unlabeled novel examples that have been classified (by an initial classifier) with high confidence in order to self-train the model \cite{liu2018learning,li2019learning,chen2019block} or fine-tuning the model on unlabeled novel examples with an auxiliary loss serving as a regularizer \cite{dhillon2019baseline,rodriguez2020embedding,lichtenstein2020tafssl}. These methods still focus on \textbf{improving} the generalization of inductive bias borrowed from the supervised learning of base classes. 

In comparison, our key motivation is that, unlabeled examples from novel classes not only can fine-tune or retrain a pre-trained model, but also can effectively train a new model from scratch. The advantage of doing so is that the model can generalize better on novel classes.
In this paper, we demonstrate the effectiveness of \textbf{an extremely simple baseline} for transductive few-shot learning.
\textbf{Our baseline does not use any labels on the base classes}. 
{We conduct self-supervised learning on unlabeled data from both the base and the novel classes to learn a feature embedding. When doing few-shot classification, we directly learn a linear classifier on top of the feature embedding from the few given labeled examples and then classify the testing examples.}
Surprisingly, this baseline significantly outperforms state-of-the-art transductive few-shot learning methods, which have additional access to base-class labels. 

The empirical performance of this baseline should not be ``the final solution'' for few-shot learning. We believe that meta-learning, metric learning, data augmentation, and transfer learning are also critical for effective few-shot learning. However, this baseline can help us interpret existing results and indicates that using self-supervised learning to learn a generalized representation could be another important tool in addressing few-shot learning. 

To investigate the best possible way to use self-supervised learning in few-shot learning, it is necessary to examine more carefully the role of features learned through self-supervision in few-shot learning. For brevity, we refer to these features as `self-supervised features'. 
\textbf{(1)} In a non-transductive few-shot learning setting, we explore the \textit{complementarity} and \textit{transferability} of supervised and self-supervised features. By directly concatenating self-supervised and supervised features, we get a 2-3\% performance boost and achieve new state-of-the-art results. We conduct cross-domain few-shot learning and show that supervised features have better transferability than self-supervised features. However, when more novel labeled examples are given, self-supervised features overtake supervised features.
\textbf{(2)} In a transductive few-shot learning setting, we show that simple off-the-shelf self-supervised learning significantly outperforms other competitors who have additional access to base-class labels. 
We confirm the performance gain is not from a better representation but from a representation that better generalizes on the novel classes. The proof is that the self-supervised features achieve the top performance on the novel classes but \emph{not} on other unseen classes.
\textbf{(3)} For both non-transductive and transductive settings, we conduct comprehensive experiments to explore the effect of different backbone architectures and datasets. We report results using a shallow ResNet, a very deep ResNet, a very wide ResNet, and a specially designed shallow ResNet that is commonly used for few-shot learning. While deeper models generally have significantly better performance for the standard classification task on both large (e.g, ImageNet~\cite{deng2009imagenet}) and small datasets (e.g., CIFAR-10) as shown in \cite{he2015deep}, the performance gain is relatively small for supervised features in few-shot learning. In comparison, self-supervised features show a much larger improvement when using a deeper network, especially in the transductive setting.
We also conduct experiments on various datasets, including large datasets, small datasets, and datasets that have small or large domain differences between base and novel classes. We show the efficiency and robustness of self-supervised features on all kinds of datasets except for very small datasets.

\secvspace
\section{Related Work}
\secvspace

\textbf{Few-shot Learning.} Few-shot learning is a classic problem \cite{miller2000learning}, which refers to learning from one or a few labeled examples for each novel class. Existing FSL methods can be broadly grouped into three categories: data augmentation, meta-learning, and metric learning. Data augmentation methods synthesize \cite{imaginaryData,Delta-encoder,chen2019multi}, hallucinate \cite{2017ICCVaug} or deform \cite{chen2019image} images to generate additional examples to address the training data scarcity. Meta-learning~\cite{MAML,Sachin2017,MetaNetwork,lee2019meta} attempts to learn a parameterized mapping from limited training examples to hidden parameters that accelerate or improve the optimization procedure. Metric learning \cite{relation_net,bateni2020improved,li2020boosting} aims at learning a transferable metric space (or embedding). MatchingNet \cite{matchingnet_1shot} and ProtoNet \cite{prototype_network} adopt cosine and Euclidean distance to separate instances belonging to different classes. Recently, some works \cite{chen2019multi,liu2020negative,tian2020rethinking} showed that learning a classifier on top of supervised features can achieve surprisingly competitive performance.

\begin{figure*}[t]
  \centering
  \includegraphics[width=16cm]{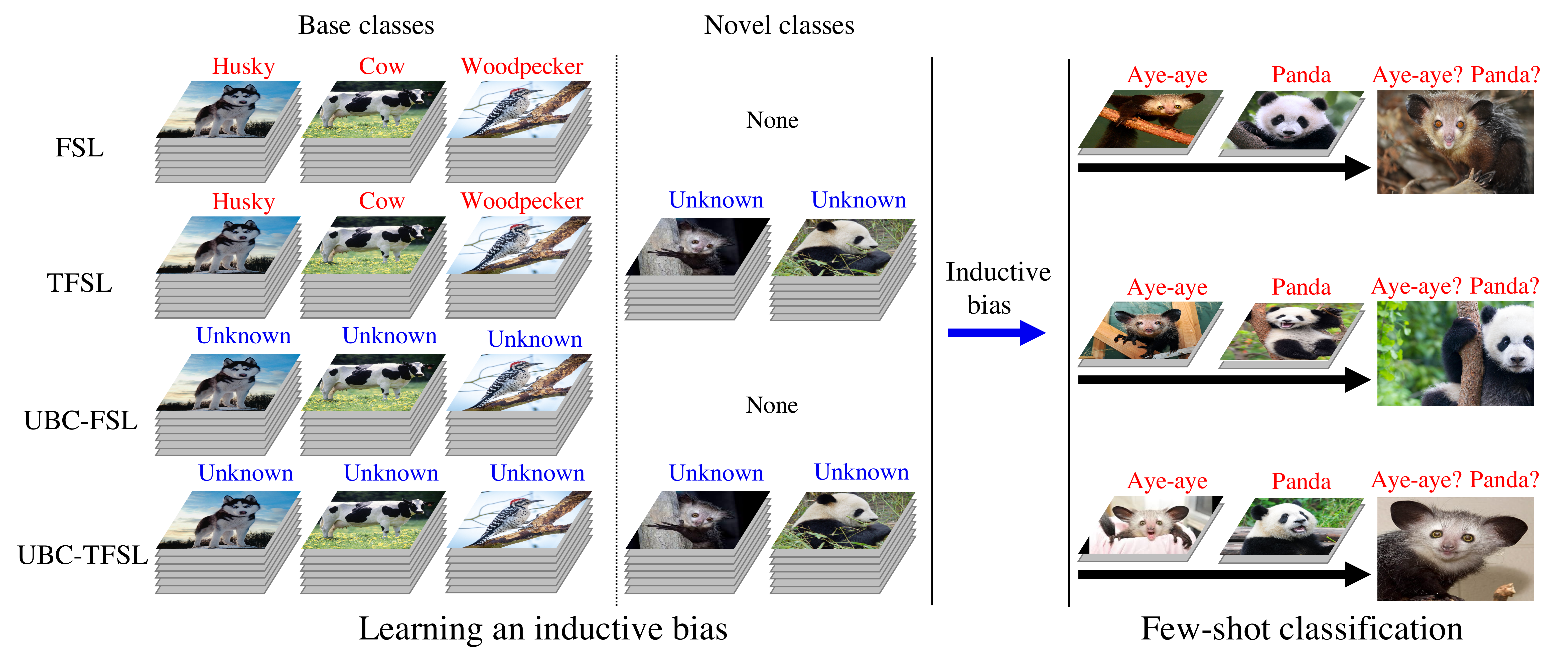}
  \captionvspace
  \vspace{-0.2cm}
  \caption{\textbf{An illustration of different few-shot learning settings.} There are four few-shot settings, including few-shot learning (FSL), transductive few-shot learning (TFSL), unlabeled-base-class few-shot learning (UBC-FSL), and unlabeled-base-class transductive few-shot learning (UBC-TFSL).
  The differences between these settings are \textbf{whether they have labels for examples from the base classes and unlabeled examples from the novel classes.} }
  \label{fig:setting}
  \figvspace
\end{figure*}

\textbf{Transductive Few-shot Learning.}
TFSL methods use the distribution support of unlabeled novel instances to help few-shot learning. Some TFSL methods \cite{liu2018learning,li2019learning,Wang_2020_CVPR} exploit unlabeled instances with high confidence to train the model. 
\cite{chen2019block} propose a data augmentation method to directly mix base examples and selected novel examples in the image domain to learn generalized features.
In addition, previous work\cite{dhillon2019baseline,rodriguez2020embedding,lichtenstein2020tafssl} seek to take unlabeled testing instances to acquire an auxiliary loss serving as a regularizer to adapt the inductive bias. 
These methods borrow inductive bias from the supervised learning of the base classes and further utilize unlabeled novel examples to improve it. 
In comparison, we show that unlabeled novel examples in addition to labeled examples of the base classes can directly develop a very strong inductive bias. 



\textbf{Self-supervised Learning.} Self-supervised learning aims to explore the internal data distribution and learns discriminative features without annotations. Some work takes predicting rotation \cite{gidaris2018unsupervised}, counting \cite{noroozi2017representation}, predicting the relative position of patches \cite{doersch2015unsupervised}, colorization \cite{zhang2016colorful,larsson2017colorization}, and solving jigsaw puzzles \cite{noroozi2016unsupervised} as self-supervised tasks to learn representations. Recently, instance discrimination \cite{wu2018unsupervised,chen2020simclr,grill2020bootstrap,tian2019contrastive} has attracted much attention. 
\cite{he2020momentum} propose a momentum contrast to update models and shows superior performance to supervised learning.
\zt{In this work, we explore the generalization ability of self-supervised features to new classes in the few-shot setting, i.e., in circumstances where few labeled examples of novel classes are given.} \e{Other works that have explored transductive techniques, e.g., \cite{he2020momentum}, have used large training sets for new classes.} 
Gidaris et al. \cite{gidaris2019boosting} and Su et al. \cite{su2019does} take rotation prediction, solving jigsaw as auxilary tasks to learn better representation on base classes to help few-shot learning. 
Tian et al. \cite{tian2020rethinking} utilize contrastive learning to learn features for non-transductive few-shot learning.
In comparison, while previous works only conduct self-supervised learning under the non-transductive, we confirm the effectiveness of self-supervised learning in a transductive few-shot setting. We claim this as our major contribution.

\secvspace
\section{Methods}
\secvspace


\label{setting}

In Fig.~\ref{fig:setting}, we illustrate our few-shot learning settings.
We denote the base category set as $C_{base}$ and the novel category set as $C_{novel}$, in which $C_{base} \cap C_{novel} = \emptyset$. Correspondingly, we denote the labeled base dataset as $D_{base} = \{(I_i,y_i)\},y_i\in C_{base}$, the labeled novel dataset as $D_{novel}=\{(I_i,y_i)\},y_i\in C_{novel}$, the unlabeled base dataset as $U_{base} = \{(I_i)\},y_i\in C_{base}$, and the unlabeled novel dataset as $U_{novel}=\{(I_i)\},y_i\in C_{novel}$. 

In a standard few-shot learning task, we are only given labeled examples from base classes so the training set is $D_{FSL} = D_{base}$. For transductive few-shot learning (TFSL), we are given $D_{TFSL} = D_{base} \cup U_{novel}$. For unlabeled-base-class few-shot learning (UBC-FSL), we have $D_{UBC-FSL} = U_{base}$. For unlabeled-base-class transductive few-shot learning (UBC-TFSL), we denote the training set as $D_{UBC-TFSL} = U_{base} \cup U_{novel}$. 
Note that UBC-TFSL has strictly less supervision than TFSL and this setting has not been explored before. We claim \textbf{we are the first to explore this setting}.

These four few-shot learning settings use the same evaluation protocol as in previous works \cite{matchingnet_1shot}. At inference time, we are given a collection of \emph{N-way-m-shot} classification tasks sampled from $D_{novel}$ to evaluate our methods.


\begin{figure*}[t]
  \centering
  \includegraphics[width=16cm]{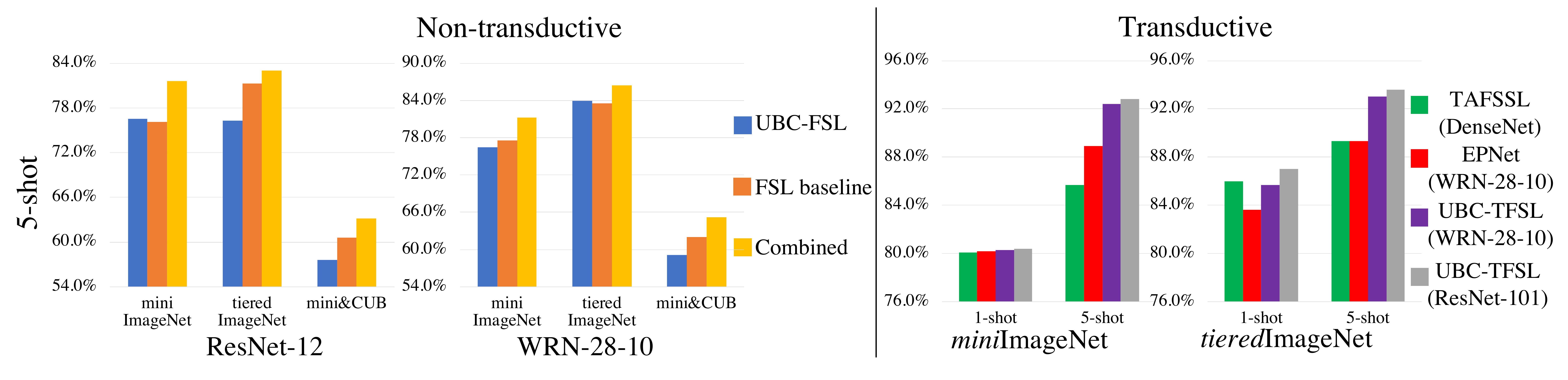}
  \captionvspace
  \vspace{-0.2cm}
  \caption{\textbf{Comparison between different methods under the non-transductive and transductive few-shot setting.} For the non-transductive few-shot learning, there is great complementarity among supervised and self-supervised features. For the transductive few-shot learning, our UBC-TFSL outperform other competitors even without using base-classes labels.
  }
  \label{fig:keyvis}
  \figvspace
\end{figure*}

\subsecvspace
\subsection{Self-supervised learning}
\subsecvspace

Here we take instance discrimination as our self-supervision task due to its efficiency.
We follow momentum contrast \cite{he2020momentum}, where each training example $x_i$ is augmented twice into $x_i^q$ and $x_i^k$. $x_i^q$ and $x_i^k$ are then fed into two encoders forming two embeddings $q_i=f_q(x_i^q)$, and $k_i=f_k(x_i^k)$. A standard log-softmax function is used to discriminate a positive pair (2 instances augmented from one image) from several negative pairs (2 instances augmented from  2 images):
\begin{equation}
L(q_i,k_i) = -\log \left ( \frac{\exp(q_i^T k_i/\tau)}{\exp(q_i^T k_i/\tau) + \sum_{j\neq i} \exp(q_i^T k_j /\tau)}\right)
\label{eqn:loss}
\end{equation}
where $\tau$ is a temperature hyper-parameter. Since our implementations are based on MoCo-v2 \cite{chen2020mocov2}, please refer to it for further details. We also try other self-supervised methods in \cref{selfmethods}.

\subsecvspace
\subsection{Evaluation Protocols}
\subsecvspace

Here we introduce our protocols for the four different few-shot learning settings. 
All protocols consist of a training phase and an evaluation phase. In the training phase, we learn a feature embedding on the training sets $D_{FSL}$, $D_{TFSL}$, $D_{UBC-FSL}$, and $D_{UBC-TFSL}$. In the evaluation phase, we evaluate the few-shot classification performance.
For simplicity and efficiency, we learn a logistic regression classifier on top of the learned feature embedding of $N*m$ training examples and then classify the testing examples. 
Training and testing examples come from the given \emph{N-way-m-shot} classification task. Such procedures are repeated $1000$ times and we report the average few-shot classification accuracies with 95\% confidence intervals. Now, we would like to introduce our methods.

\textbf{Few-shot learning baseline.} 
We learn our embedding network on $D_{FSL}$ using cross-entropy loss under a standard classification process. We use the logit layer as the feature embedding as it is slightly better than the pre-classification layer. This baseline is very simple and achieve the state-of-the-art performance. 


\textbf{Unlabeled-base-class few-shot learning.} For UBC-FSL, we learn from self-supervised supervision on $D_{UBC-FSL}$. We follow MoCo-v2 to do instance discrimination. The output of the final layer of the model is used as the feature embedding. 

\textbf{Unlabeled-base-class transductive few-shot learning.} For UBC-TFSL, our method is similar to our UBC-FSL method. The difference is that we train on $D_{UBC-TFSL}$, which has additional access to unlabeled test instances. 

\textbf{Combination of FSL baseline and UBC-FSL.} This method works under standard, non-transductive, few-shot learning setting. We explore the complementarity between supervised features (from the FSL baseline) and self-supervised features (from UBC-FSL). We directly concatenate normalized supervised features and normalized self-supervised features and then do normalization again. T his feature is used as the feature embedding and we refer this method as ``Combined''.


\begin{table*}[h]
\caption{\textbf{Top-1 accuracies(\%) on \emph{mini}ImageNet and \emph{tiered}ImageNet.} We report the mean of 1000 randomly generated test episodes as well as the 95\% confidence intervals.  The top results are highlighted in \first{blue} and the second-best results in \second{green}.
We provide results on Caltech-256 and \emph{mini}ImageNet\&CUB in the \textbf{supplementary}.
}
    \centering
    \small
\begin{tabular}{cclcccc}
\hline 
 &  &  & \multicolumn{2}{c}{
\textbf{\emph{mini}ImageNet}
} & \multicolumn{2}{c}{\textbf{\emph{tiered}ImageNet}}\tabularnewline
\textbf{setting} & \textbf{method} & \textbf{backbone} & \textbf{1-shot} & \textbf{5-shot} & \textbf{1-shot} & \textbf{5-shot}\tabularnewline
\hline 
\multirow{17}{*}{\textbf{Non-transductive}} 
  
 & MetaOptNet & ResNet-12$^*$ & 62.6$\pm$0.6 & 78.6$\pm$0.4 & 65.9$\pm$0.7 & 81.5$\pm$0.5\tabularnewline
 
 & Distill & ResNet-12$^*$ & \second{64.8$\pm$0.6} & \first{82.1$\pm$0.4} & 71.5$\pm$0.6 & 86.0$\pm$0.4\tabularnewline
 
 
 
 
 & Neg-Cosine & ResNet-12$^*$ & 63.8$\pm$0.8 & 81.5$\pm$0.5 & - & -\tabularnewline
 
 & Neg-Cosine & WRN-28-10 & 61.7$\pm$0.8 & 81.7$\pm$0.5 & - & -\tabularnewline
 

 & UBC-FSL (Ours) & ResNet-12$^*$ & 47.8$\pm$0.6 & 68.5$\pm$0.5 & 52.8$\pm$0.6 & 69.8$\pm$0.6\tabularnewline
 
 & UBC-FSL (Ours) & ResNet-12 & 56.9$\pm$0.6 & 76.5$\pm$0.4 & 58.0$\pm$0.7 & 76.3$\pm$0.5\tabularnewline
 
 & UBC-FSL (Ours) & ResNet-50 & 56.2$\pm$0.6 & 75.4$\pm$0.4 & 66.6$\pm$0.7 & 83.1$\pm$0.5\tabularnewline
 
 & UBC-FSL (Ours) & ResNet-101 & 57.5$\pm$0.6 & {77.2$\pm$0.4} & {68.0$\pm$0.7} & {84.3$\pm$0.5}\tabularnewline

 & UBC-FSL (Ours) & WRN-28-10 & 57.1$\pm$0.6 & {76.5$\pm$0.4} & {67.5$\pm$0.7} & {83.9$\pm$0.5}\tabularnewline
 
 & FSL baseline & ResNet-12$^*$ & 61.7$\pm$0.7 & 79.4$\pm$0.5 & 69.6$\pm$0.7 & 84.2$\pm$0.6\tabularnewline
 
 & FSL baseline & ResNet-12 & 61.1$\pm$0.6 & 76.1$\pm$0.6 & 66.4$\pm$0.7 & 81.3$\pm$0.5\tabularnewline
  
 & FSL baseline & ResNet-50 & 61.3$\pm$0.6 & 76.0$\pm$0.4 & 69.4$\pm$0.7 & 83.3$\pm$0.5\tabularnewline
 
 & FSL baseline & ResNet-101 & 62.7$\pm$0.7 & 77.6$\pm$0.5 & 70.5$\pm$0.7 & 83.8$\pm$0.5\tabularnewline
 
 & FSL baseline & WRN-28-10 & 62.4$\pm$0.7 & 77.5$\pm$0.5 & 70.2$\pm$0.7 & 83.5$\pm$0.5\tabularnewline
 
  & Combined (Ours) & ResNet-12$^*$ & 59.8$\pm$0.8 & 73.3$\pm$0.7 & 69.2$\pm$0.7 & 82.0$\pm$0.6\tabularnewline
 
 & Combined (Ours) & ResNet-12 & 63.8$\pm$0.7 & 79.9$\pm$0.6 & 67.8$\pm$0.7 & 83.0$\pm$0.5\tabularnewline

 & Combined (Ours) & ResNet-50 & 63.9$\pm$0.9 & 79.9$\pm$0.5 & {72.3$\pm$0.7} & {86.1$\pm$0.5}\tabularnewline
 
 & Combined (Ours) & ResNet-101 & \first{65.6$\pm$0.6} & \second{81.6$\pm$0.4} & \first{73.5$\pm$0.7} & \first{86.7$\pm$0.5}\tabularnewline
 
  & Combined (Ours) & WRN-28-10 & {65.2$\pm$0.6} & {81.2$\pm$0.4} & \second{73.1$\pm$0.7} & \second{86.4$\pm$0.5}\tabularnewline
\hline 

\multirow{11}{*}{\textbf{Transductive}} & ICI & ResNet-12$^*$ & 66.8$\pm$1.1 & 79.1$\pm$0.7 & 80.7 $\pm$1.1 & 87.9$\pm$0.6\tabularnewline

 & ICI & ResNet50 & 60.2$\pm$1.1 & 75.2$\pm$0.7 & 78.6$\pm$1.1 & 86.8$\pm$0.6\tabularnewline
 
 & ICI & ResNet101 & 64.3$\pm$1.2 & 78.1$\pm$0.7 & 82.4$\pm$1.0 & 89.4$\pm$0.6\tabularnewline
 
 & TAFSSL & DenseNet& 80.1$\pm$0.2 & 85.7$\pm$0.1 & {86.0$\pm$0.2} & 89.3$\pm$0.1\tabularnewline
 
 & EPNet & WRN-28-10 & 79.2$\pm$0.9 & 88.0$\pm$0.5 & 83.6$\pm$0.9 & 89.3$\pm$0.5\tabularnewline
 
  & EPNet (full) & WRN-28-10 & {80.2$\pm$0.8} & {88.9$\pm$0.5} & 84.8$\pm$0.8 & 89.9$\pm$0.6\tabularnewline

 & UBC-TFSL (Ours) & ResNet-12$^*$ & 51.1$\pm$0.9 & 74.6$\pm$0.6 & 57.2$\pm$0.6 & 74.7$\pm$0.6\tabularnewline
 
 & UBC-TFSL (Ours) & ResNet-12 & 70.3$\pm$0.6 & 86.9$\pm$0.3 & 65.7$\pm$0.7 & 81.4$\pm$0.5\tabularnewline
 
 & UBC-TFSL (Ours) & ResNet-50 & 79.1$\pm$0.6 & {92.1$\pm$0.3} & 81.0$\pm$0.6 & {90.7$\pm$0.4}\tabularnewline
 
 & UBC-TFSL (Ours) & ResNet-101 & \first{80.4$\pm$0.6} & \first{92.8$\pm$0.2} & \first{87.0$\pm$0.6} & \first{93.6$\pm$0.3}\tabularnewline
 
 & UBC-TFSL (Ours) & WRN-28-10 & \second{80.3$\pm$0.6} & \second{92.4$\pm$0.2} & \second{85.7$\pm$0.6} & \second{93.0$\pm$0.3}\tabularnewline
\hline 
\end{tabular}

\label{tab:benchmark}
\figvspace
\end{table*}

\secvspace
\section{Experiments}
\secvspace


We define two types of experiments based upon whether the base and novel classes come from the same dataset or not. We refer to the standard FSL paradigm in which the base and novel classes come from the same dataset (e.g., ImageNet) as \textit{single-domain} FSL. We also perform experiments in which the novel classes are chosen from a separate dataset, which we call \textit{cross-domain} FSL. In cross-domain FSL, the domain differences between the base and novel classes are much larger than the single-domain FSL. For both setting, we report 5-way-1-shot and 5-way-5-shot accuracies.

\textbf{Datasets.} For single-domain FSL, we run experiments on three datasets: \emph{mini}ImageNet~\cite{matchingnet_1shot}, \emph{tiered}ImageNet~\cite{ren2018meta}, and Caltech-256~\cite{griffin2007caltech}. 
The \emph{mini}ImageNet contains 100 classes randomly selected from ImageNet~\cite{deng2009imagenet} with 600 images per class. We follow \cite{Sachin2017} to split the categories into 64 base, 16 validation, and 20 novel classes. The \emph{tiered}ImageNet is another subset of ImageNet but has far more classes (608 classes). These classes are first divided into 34 groups and then further divided into 20 training groups (351 classes), 6 validation groups (97 classes), and 8 testing groups (160 classes), which ensure the distinction between training and testing sets. Caltech-256 (Caltech) has 30607 images from 256 classes. Following \cite{chen2019multi}, we split it into 150, 56, and 50 classes for training, validation, 
and testing respectively. 

 
For the cross-domain experiments, we construct a dataset that has high dissimilarity between base and novel classes by drawing the base classes from one dataset and the novel classes from another. 
We denote this dataset as '\emph{mini}ImageNet\&CUB', which is a combination of \emph{mini}ImageNet and CUB-200-2011 (CUB) dataset \cite{wah2011caltech}. 
CUB is a fine-grained image classification dataset including 200 bird classes and 11788 bird images. We follow \cite{2018arXiv180204376H} to split the categories into 100 base, 50 validation, and 50 novel classes. 
In \emph{mini}ImageNet\&CUB, the training set (base classes) contains 64 classes from \emph{mini}ImageNet and the testing set (novel classes) contains 100 classes from CUB. Specifically, the 64 classes in the training set are the 64 base classes in \emph{mini}ImageNet and the 100 classes in the test set are the 100 base classes in CUB.

\textbf{Competitors.} We compare our methods with the top few-shot learning methods: MetaOptNet \cite{lee2019meta}, Distill \cite{tian2020rethinking}, and Neg-Cosine \cite{liu2020negative}.
We also compare with three transductive few-shot learning methods:
ICI \cite{Wang_2020_CVPR}, TAFSSL \cite{lichtenstein2020tafssl}, and EPNet \cite{rodriguez2020embedding}. TFSL methods have 100 unlabeled images per novel class by default. EPNet (full)
and our UBC-TFSL uses all of the images of novel classes as unlabeled training samples.

\textbf{Implementation details.} 
Most of our settings are the same as \cite{chen2020mocov2}. We use a mini-batch size of 256 with 8 GPUs. We set the learning rate as 0.03 and use cosine annealing to decrese the learning rate. The feature dimension for contrastive loss is 128. The momentum for memory update is 0.5 and the temperature is set as 0.07. For \emph{mini}ImageNet, \emph{mini}ImageNet\&CUB, and Caltech-256, we sample 2048 negative pairs in our contrastive loss and train 1000 epochs. For \emph{tiered}ImageNet, we sample 20480 negative pairs and train 800 epochs.

\textbf{Architecture.} We use ResNet-12$^*$, ResNet-12, ResNet-50, ResNet-101, and WRN-28-10 as our backbone architecture. ResNet-12$^*$ is a modified version of ResNet-12 and will be introduced in Sec.\cref{arch}. WRN-28-10~\cite{BMVC2016_87} is a very wide version of ResNet-10 and have 36.5M parameters whereas ResNet-50 and ResNet-101 have 25.6M and 44.4M parameters respectively.

\subsecvspace
\subsection{\zt{Self-supervised learning can develop a strong inductive bias with no base-class labels} }
\subsecvspace
\label{alone}
\label{benchmark}

\begin{figure*}[t]
  \centering
  \includegraphics[width=16cm]{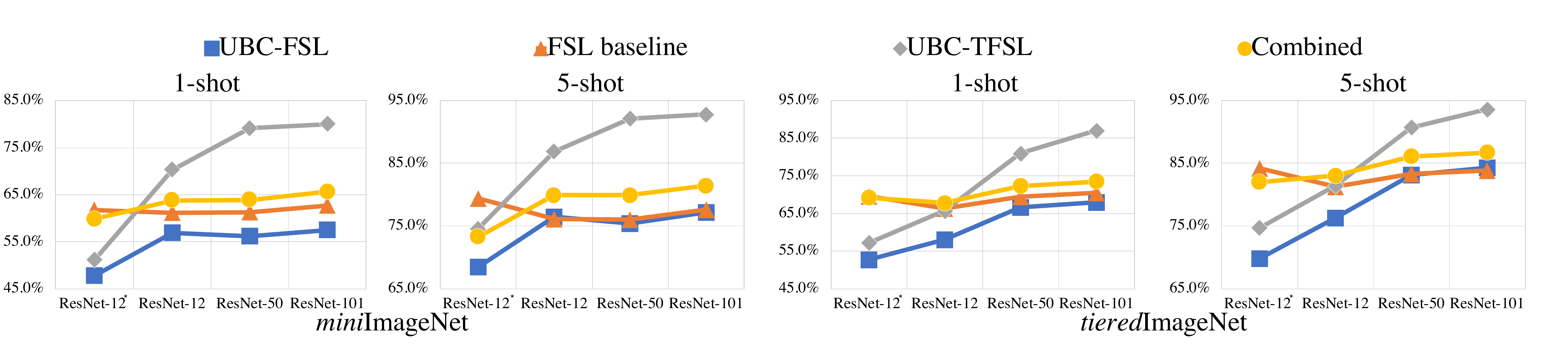}
  \captionvspace
  \vspace{-0.2cm}
  \caption{\textbf{Few-shot classification accuracy with various depths of backbone architectures. } Our UBC-FSL, FSL baseline, UBC-TFSL, and Combined have better performance with a deeper network. 
  \textbf{The performance gain is relatively small for supervised features (FSL baseline) and large for self-supervised features (UBC-FSL), especially in a transductive setting (UBC-TFSL).}
  }
  \label{fig:arch}
  \figvspace
\end{figure*}

\begin{figure*}[t]
  \centering
  \includegraphics[width=16cm]{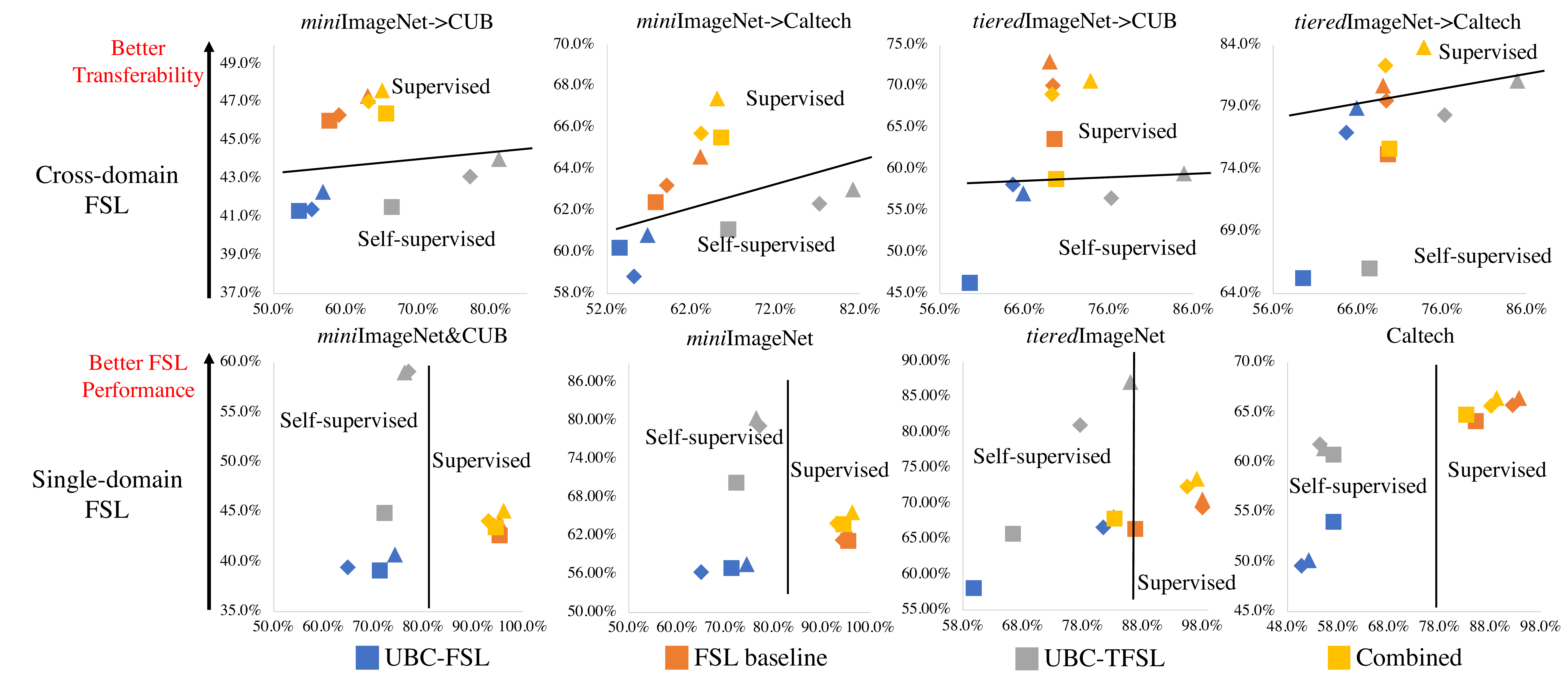}
  \captionvspace
  \vspace{-0.2cm}
  \caption{ 
  \textbf{Accuracy of 1-shot cross-domain FSL  (first row) or single-domain FSL (second row).} First row: we visualize 1-shot test accuracy on the source dataset (x-axis) and the target dataset (y-axis). Second row: we visualize 1-shot accuracy on the base classes (x-axis) and the novel classes (y-axis). Squares, diamonds, and triangles denote ResNet-12, ResNet-50, and ResNet-101 respectively. We provide detailed statistics in the supplementary. 
  From the first row, the results suggest that \textbf{supervised features are better when transferring to a new target dataset} even if self-supervised features (UBC-TFSL) have more training data and better performance on the source dataset. From the second row, the results suggest that in few-shot learning, even if supervised features have better performance on base classes, it underperforms self-supervised features (UBC-TFSL) on novel classes. It confirms that \textbf{UBC-TFSL benefit from the representation that better generalized on the novel classes}. 
}
  \label{fig:transfer}
  \figvspace
\end{figure*}

\begin{figure*}[t]
  \centering
  \includegraphics[width=13cm]{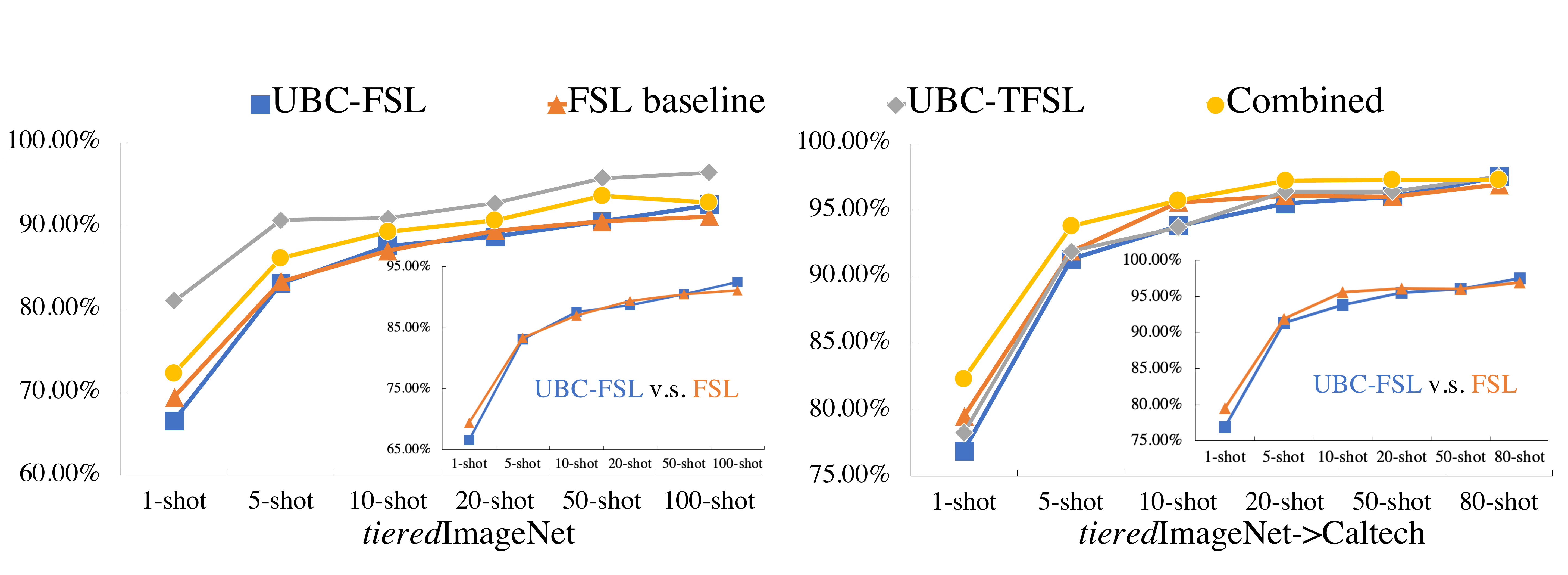}
  \captionvspace
  \vspace{-0.2cm}
  \caption{\textbf{Few-shot classification accuracy with larger shots.} We use ResNet-50 as our backbone architecture and evaluate on \emph{tiered}ImageNet and Caltech (transferred from \emph{tiered}ImageNet). Training on same data, supervised features (FSL baseline) outperform self-supervised features (UBC-FSL) in few-shot setting. However, self-supervised features are better when large numbers (100-shot) of labeled novel examples are given.  }
  \label{fig:shots}
  \figvspace
\end{figure*}

\cite{su2019does} shed light on improving few-shot learning with self-supervision and claim that ``Self-supervision alone is not enough'' for FSL. We agree there is still a gap between unlabeled-base-class  few-shot  learning and few-shot learning.
However, in the transductive few-shot classification setting, we present the surprising result that \textbf{state-of-the-art performance can be obtained without using any labeled examples from the base classes at all.} 

The results on \emph{mini}ImageNet and \emph{tiered}ImageNet are shown in Table \ref{tab:benchmark}. A better visualization is shown in Fig.~\ref{fig:keyvis}. Results on Caltech-256 and \emph{mini}ImageNet\&CUB are provided in supplementary material.
We notice that \textbf{(1) UBC-FSL shows some potential.} Even without any base-class labels, it only underperforms the state-of-the-art few-shot methods by $2-7\%$ in 1-shot and 5-shot accuracy on \emph{mini}ImageNet and \emph{tiered}ImageNet. 
\textbf{(2) There is great complementarity among supervised features and self-supervised features.} 
Combining supervised and self-supervised features (``Combined'') beats the FSL baseline on all four datasets for all backbone networks. Specifically, it gives 4\% and 2.9\% improvements in 5-shot accuracy on \emph{mini}ImageNet and \emph{tiered}ImageNet when using ResNet-101. Also, it beats all other FSL competitors on \emph{tiered}ImageNet.
\textbf{(3) For the transductive few-shot classification setting, state-of-the-art can be obtained without actually using any labeled examples at all.} Even without any base-class labels, UBC-TFSL significantly surpasses all other methods. 
In Table~\ref{tab:benchmark}, it outperforms all other TFSL methods by $3.5\%$ and $3.9\%$ for 5-shot accuracy on \emph{mini}ImageNet and \emph{tiered}ImageNet respectively. 
\textbf{(4) The FSL baseline struggles to learn a strong inductive bias with high dissimilarity between base and novel classes (cross-domain) whereas such dissimilarity has a relatively minor effect on UBC-TFSL.} In \emph{mini}ImageNet\&CUB (please refer to supplementary), UBC-TFSL outperforms the FSL baseline by~$15\%$ and ~$13\%$ for 1-shot and 5-shot accuracy respectively.

\subsecvspace
\subsection{A deeper network is better}
\subsecvspace
\label{arch}

While deeper models generally have better performance for the standard classification task on both large (e.g, ImageNet \cite{deng2009imagenet}) and small dataset(e.g., CIFAR-10) as shown in \cite{he2015deep}, most previous few-shot methods \cite{tian2020rethinking,lee2019meta} report the best results with a modified version (ResNet-12$^*$) of ResNet-12\cite{he2015deep}.
ResNet-12$^*$ employs several modifications, including making it $1.25\times$ wider, changing the input size from $224\times 224$ to $84\times 84$, using Leaky ReLU's instead of ReLU's, adding additional Dropblock layers \cite{ghiasi2018dropblock}, and removing the global pooling layer after the last residual block. 
We feel that the effect of different backbone architecture is not very clear in few-shot learning literature. We want to know if using a very deep network (e.g., ResNet-101) can bring significant improvement in few-shot classification as in the standard classification task.
More importantly, we want to explore the differences between supervised and self-supervised features when using various backbone architectures in few-shot learning.

As shown in Table \ref{tab:benchmark}, we report results using ResNet-12$^*$, ResNet-12, ResNet-50, ResNet-101, and WRN-28-10. To better compare the effect of depth of backbone architecture, we visualize the performance in Fig.~\ref{fig:arch}. We notice that (1) ResNet-101 have the best performance and all our baselines benefit from deeper network in most cases. (2) The commonly used ResNet-12$^*$ work well for FSL baseline but do not suit for self-supervised learning based baselines. (3) The wide network WRN-28-10 has very good performance on all our baselines, only slightly underperform ResNet-101.  We confirm that few-shot learning can actually benefit from a deeper or wider backbone architecture. \textbf{The performance gain is small for supervised features (FSL baseline) and large for self-supervised features (UBC-FSL), especially in a transductive setting (UBC-TFSL).} 
\begin{figure*}[h]
  \centering
  \includegraphics[width=13cm]{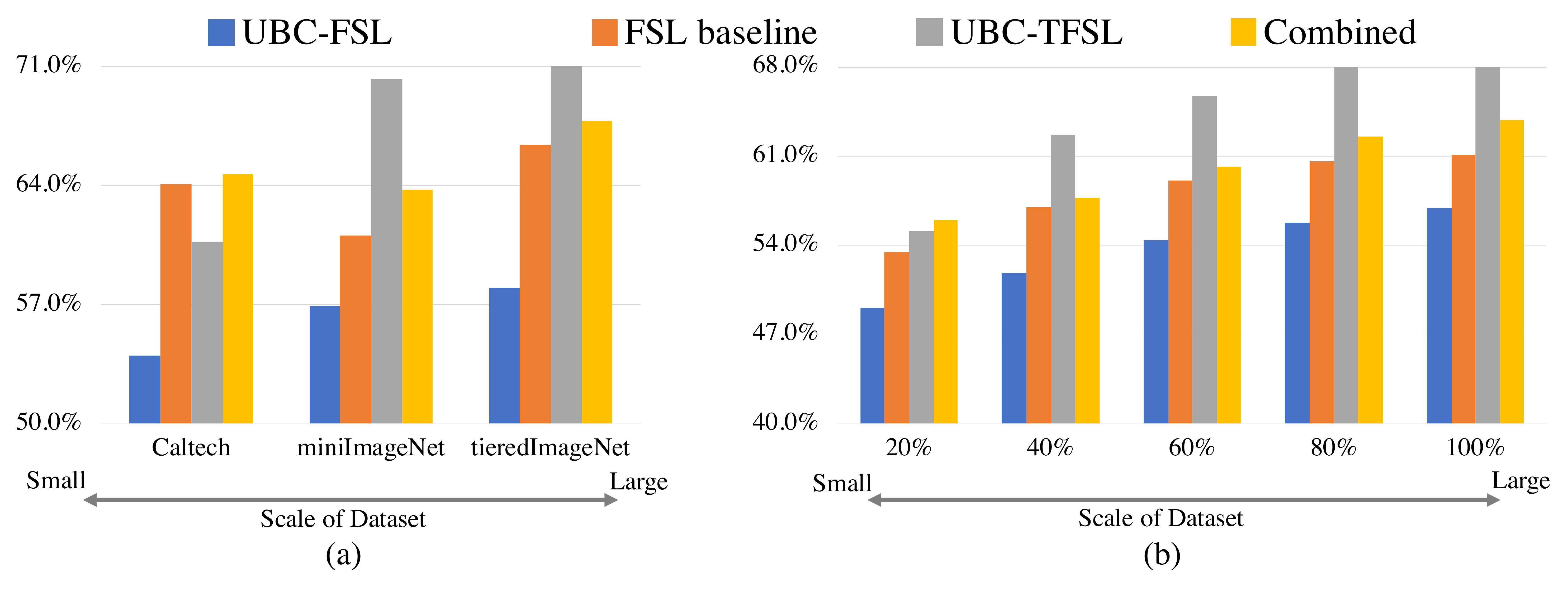}
  \captionvspace
  \vspace{-0.2cm}
  \caption{\textbf{1-shot testing accuracy under various scales of dataset size.} ResNet-12 is our backbone architecture. In (a), we compare UBC-FSL, FSL baseline, UBC-TFSL, and Combined on three datasets of different sizes (30607, 60000, and 779165 images). In (b), we randomly select part of \emph{mini}ImageNet (e.g., 20\% of the whole dataset) and compare our methods.}
  \label{fig:scale}
  \figvspace
\end{figure*}

\subsecvspace
\subsection{Supervised vs.~self-supervised features in cross-domain FSL}
\subsecvspace

Another interesting question is whether models learned in a single domain can perform well in a new domain (with highly dissimilar classes). To study this, we conduct cross-domain FSL, in which we learn models on \emph{mini}ImageNet or \emph{tiered}ImageNet and evaluate our models on Caltech-256 and CUB. Specifically, the FSL baseline and UBC-FSL are trained on base classes of the source dataset, and UBC-TFSL are trained on both base and novel classes of the source dataset. Then, we evaluate our methods on the testing set of target datasets (Caltech-256 and CUB). 

Notice that the way we are applying the UBC-TFSL model, it does not qualify as a true transductive setting, since the model does not have access to unlabeled data from the testing set. Instead, we are testing whether this model can improve its performance on cross-domain classes with unlabeled data from \textbf{additional} classes in the source data set. 

Previous work \cite{he2020momentum} compares supervised  and self-supervised features when transferring to a new domain for classification, object detection, and instance segmentation. It shows that self-supervised features have better transferability for these tasks. However, the conclusion is based on when large numbers of labeled examples are used to learn the final linear classifier. In few-shot setting, we show that \textbf{supervised features do better than self-supervised features}.


In the first row of Fig.~\ref{fig:transfer}, we compare UBC-FSL, FSL baseline, UBC-TFSL, and Combined in cross-domain FSL. The x-axis and y-axis denote the 1-shot testing accuracy on the source and target dataset respectively. Surprisingly, supervised features (FSL baseline, Combined) significantly outperform self-supervised features (UBC-FSL, UBC-TFSL) on the target dataset even if they have lower accuracy on the source dataset. In the second row of Fig.~\ref{fig:transfer}, we visualize the performance of our methods on base and novel classes in single-domain FSL. The x-axis and y-axis denote the 1-shot accuracy on base and novel classes respectively. As you can see, UBC-TFSL (gray points) outperforms FSL baseline (orange) on novel classes but underperforms on base classes.
These experiments show that UBC-TFSL has mediocre performance when it does \textbf{not} have access to unlabeled data from the test classes, but performs extremely well when it does. In other words, it is not simply access to additional unlabeled data that helps, but rather, data from the test classes themselves.


\subsecvspace
\subsection{Supervised vs.~self-supervised features with larger shots}
\subsecvspace

\label{shots}

In Fig.~\ref{fig:shots}, we compare UBC-FSL, the FSL baseline, UBC-TFSL and Combined with larger shots using ResNet-50 on \emph{tiered}ImageNet and \emph{tiered}ImageNet-Caltech (cross-domain FSL). For 1-shot learning, there is a large gap around 5\% between UBC-FSL and the FSL baseline. However, as the shots become larger, this gap gradually diminishes. For 100-shot on \emph{tiered}ImageNet and 80-shot on Caltech, UBC-FSL even outperforms the FSL baseline by 1.3\% and
0.6\% respectively. 

We suggest that \textbf{supervised features may contain higher-level semantic concepts that are easier to digest with a few training instances} while {self-supervised features have better transferability with abundant training data}. This statement is compatible with previous work~\cite{he2020momentum}, which use abundant labeled data to learn the final classification layer and claims that self-supervised features have better transferability.


\subsecvspace
\subsection{Supervised vs.~self-supervised features and dataset size}
\subsecvspace
\label{size}

In this section, we compare supervised and self-supervised features under various dataset sizes. We conduct experiments on Caltech, \emph{mini}ImageNet, and \emph{tiered}ImageNet, which have 30607, 60000, and 779165 images respectively. We also randomly select subsets of \emph{mini}ImageNet (20\%, 40\%, 60\%, 80\%, and 100\%) and report the 1-shot accuracy. An equal portion of examples from each class are randomly selected.
As shown in Fig.~\ref{fig:scale}, self-supervised features (UBC-TFSL) significantly outperform other methods with a big dataset. However, when the dataset is small (e.g., Caltech-256 and 20\% of \emph{mini}ImageNet), it is overtaken by the FSL baseline.
This result suggests that \textbf{supervised features are more robust to dataset size.} 



\subsecvspace
\subsection{Comparing different self-supervised methods}
\subsecvspace
\label{selfmethods}

\begin{table}[t]
\caption{\textbf{Few-shot classification accuracy with different self-supervised methods.} We run experiments using MoCo-v2~\cite{chen2020mocov2}, CMC~\cite{tian2019contrastive}, and SimCLR~\cite{chen2020simclr} as our self-supervised methods to learn the feature embedding.
The top results are highlighted in \first{blue} and the second-best results in \second{green}.  }
    \centering
    \small
    
\begin{tabular}{llcccc}
\hline 
  & &  \multicolumn{2}{c}{
\textbf{\emph{mini}ImageNet}
} \tabularnewline
 \textbf{method} & \textbf{backbone} & \textbf{1-shot} & \textbf{5-shot} & \tabularnewline
\hline 

 UBC-FSL (MoCo-v2) & ResNet-101 & \second{57.5$\pm$0.6} & \first{77.2$\pm$0.4}\tabularnewline
 
 UBC-FSL (CMC) & ResNet-101 & 56.9$\pm$0.6 & \second{76.9$\pm$0.5}\tabularnewline
 
 UBC-FSL (SimCLR) & ResNet-101 & \first{57.6$\pm$0.7} & 76.7$\pm$0.6\tabularnewline
 \hline

 UBC-TFSL (MoCo-v2) & ResNet-101 & \first{80.4$\pm$0.6} & \first{92.8$\pm$0.2}\tabularnewline
 
 UBC-TFSL (CMC) & ResNet-101 & \second{79.7$\pm$0.6} & 92.1$\pm$0.3\tabularnewline
 
 UBC-TFSL (SimCLR) & ResNet-101 & 79.5$\pm$0.7 & \second{92.2$\pm$0.3}\tabularnewline

\hline 
\end{tabular}

\label{tab:compare}
\figvspace
\vspace{-0.1cm}
\end{table}

As \zt{shown in Table \ref{tab:compare}, we compare three different instance discrimination methods to learn the feature embedding. Here we compare MoCo-v2~\cite{chen2020mocov2}, CMC~\cite{tian2019contrastive}, and SimCLR~\cite{chen2020simclr}. From the results, we can see that all these self-supervised methods can learn a powerful inductive bias, especially in the transductive setting, suggesting that most self-supervised methods can be generalized to learn a good embedding for few-shot learning. }

\secvspace
\section{Conclusion}
\secvspace

Most previous FSL methods borrow a strong inductive bias from the supervised learning of base classes. In this paper, we show that no base class labels are needed to develop such an inductive bias and that self-supervised learning can provide a powerful inductive bias for few-shot learning.
{We examine the role of features learned  through  self-supervision in few-shot learning through comprehensive experiments.}





{\small
\bibliographystyle{ieee_fullname}
\bibliography{cvprbib}
}





\newpage
\appendix

\setcounter{table}{0}
\setcounter{figure}{0}
\renewcommand{\thetable}{A\arabic{table}}
\renewcommand{\thefigure}{A\arabic{figure}}
\renewcommand{\thesubsection}{A\arabic{subsection}}
\section*{Appendix}

\subsection{Compare to TFSL using the same backbone network}
To further make a fair comparison with TFSL methods, we implement EPNet\footnote{We use public code available at https://github.com/ElementAI/embedding-propagation.} and use ResNet-12, ResNet-50, ResNet-101 as its backbone. All experiments setting are the same and we have 600 unlabeled images per novel class. As shown in Table \ref{tab:epnet}, we report results on \emph{mini}ImageNet. 

\begin{table}[h]
    \centering
    \small
    
\begin{tabular}{llcccc}
\hline 
  & &  \multicolumn{2}{c}{
\textbf{\emph{mini}ImageNet}
} \tabularnewline
 \textbf{method} & \textbf{backbone} & \textbf{1-shot} & \textbf{5-shot} & \tabularnewline
\hline 

 EPNet & ResNet-12 & {75.9$\pm$1.0} &{84.8$\pm$0.6}\tabularnewline
 
 EPNet & ResNet-50 & 75.4$\pm$1.1 & {84.3$\pm$0.7}\tabularnewline
 
 EPNet & ResNet-101 & {76.1$\pm$0.8} & 86.0$\pm$0.7\tabularnewline

 UBC-TFSL  & ResNet-12 & {70.3$\pm$0.6} & {86.9$\pm$0.3}\tabularnewline
 
 UBC-TFSL  & ResNet-50 & \second{79.1$\pm$0.6} & \second{92.1$\pm$0.3}\tabularnewline
 
 UBC-TFSL  & ResNet-101 & \first{80.4$\pm$0.6} & \first{92.8$\pm$0.2}\tabularnewline

\hline 
\end{tabular}
\caption{\textbf{Comparison between our UBC-TFSL and TFSL using the same backbone network.} We compare UBC-TFSL and TFSL methods using the same backbone network. 
We report the mean of 1000 randomly generated test episodes as well as the 95\% confidence intervals. The top results are highlighted in \first{blue} and the second-best results in \second{green}.  }
\label{tab:epnet}
\figvspace
\end{table}

\begin{table*}[htbp]
    \centering
    \small
    
\begin{tabular}{cclcccc}
\hline 
 &  &  & \multicolumn{2}{c}{
\textbf{Caltech}
} & \multicolumn{2}{c}{\textbf{\emph{mini}ImageNet\&CUB}}\tabularnewline
\textbf{setting} & \textbf{method} & \textbf{backbone} & \textbf{1-shot} & \textbf{5-shot} & \textbf{1-shot} & \textbf{5-shot}\tabularnewline
\hline 
\multirow{12}{*}{\textbf{Non-transductive}} 
 
 & UBC-FSL (Ours) & ResNet-12$^*$ & 48.7$\pm$0.6 & 68.9$\pm$0.6 & 36.0$\pm$0.5 & 54.3$\pm$0.5\tabularnewline
 
 & UBC-FSL (Ours) & ResNet-12 & 54.0$\pm$0.6 & 74.6$\pm$0.5 & 39.1$\pm$0.6 & 57.6$\pm$0.5\tabularnewline
  
 & UBC-FSL (Ours) & ResNet-50 & 49.6$\pm$0.7 & 69.0$\pm$0.5 & 39.4$\pm$0.6 & 57.7$\pm$0.5\tabularnewline
 
 & UBC-FSL (Ours) & ResNet-101 & 50.1$\pm$0.6 & 69.9$\pm$0.5 & 40.7$\pm$0.6 & 59.4$\pm$0.6\tabularnewline
 
 & FSL baseline & ResNet-12$^*$ & 65.7$\pm$0.6 & 81.5$\pm$0.5 & 42.8$\pm$0.5 & 60.9$\pm$0.6\tabularnewline
 
 & FSL baseline & ResNet-12 & 64.1$\pm$0.6 & 80.5$\pm$0.6 & 42.6$\pm$0.6 & 60.6$\pm$0.5\tabularnewline
 
 & FSL baseline & ResNet-50 & 65.7$\pm$0.7 & 81.9$\pm$0.3 & 43.6$\pm$0.6 & 62.1$\pm$0.5\tabularnewline
 
 & FSL baseline & ResNet-101 & \second{66.4$\pm$0.6} & {82.5$\pm$0.4} & {43.9$\pm$0.6} & {62.4$\pm$0.6}\tabularnewline
 
 & Combined (Ours) & ResNet-12$^*$ & 65.4$\pm$0.6 & 82.7$\pm$0.5 & 42.9$\pm$0.5 & 61.7$\pm$0.7\tabularnewline
 
 & Combined (Ours) & ResNet-12 & 64.7$\pm$0.6 & 82.4$\pm$0.4 & 43.4$\pm$0.6 & 63.2$\pm$0.5\tabularnewline

 & Combined (Ours) & ResNet-50 & 65.6$\pm$0.6 & \second{82.8$\pm$0.4} & \second{44.1$\pm$0.6} & \second{64.4$\pm$0.5}\tabularnewline
 
 & Combined (Ours) & ResNet-101 & \first{66.5$\pm$0.5} & \first{83.2$\pm$0.4} & \first{45.1$\pm$0.6} & \first{65.6$\pm$0.5}\tabularnewline
\hline 

\multirow{4}{*}{\textbf{Transductive}} & UBC-TFSL (Ours) & ResNet-12$^*$ & 56.4$\pm$0.6 & 74.8$\pm$0.6 & 39.7$\pm$0.4 & 58.9$\pm$0.5\tabularnewline
 
 & UBC-TFSL (Ours) & ResNet-12 & 60.7$\pm$0.7 & 80.0$\pm$0.5 & 44.9$\pm$0.6 & 65.0$\pm$0.6\tabularnewline
 
 & UBC-TFSL (Ours) & ResNet-50 & \first{61.8$\pm$0.6} & \first{81.4$\pm$0.5} & \first{59.1$\pm$0.8} & \first{76.2$\pm$0.6}\tabularnewline
 
 & UBC-TFSL (Ours) & ResNet-101 & \second{61.4$\pm$0.6} & \second{80.3$\pm$0.5} & \second{59.0$\pm$0.8} & \second{75.5$\pm$0.6}\tabularnewline
\hline 
\end{tabular}
\caption{\textbf{Top-1 accuracies(\%) on Caltech-256 and \emph{mini}ImageNet\&CUB.} We report the mean of 1000 randomly generated test episodes as well as the 95\% confidence intervals. The top results are highlighted in \first{blue} and the second-best results in \second{green}.}
\label{tab:Caltech}
\end{table*}

\subsection{Results on Caltech-256 and miniImageNet\&CUB}

We report our results on Caltech-256 and miniImageNet\&CUB in Table \ref{tab:Caltech}.

\subsection{Results for cross-domain FSL}
\label{crossFSL}

\begin{table*}[htbp]
    \centering
    \small
    
\begin{tabular}{clcccc}
\hline 
  & &  \multicolumn{2}{c}{
\textbf{\emph{mini}ImageNet$\rightarrow$Caltech}
} & \multicolumn{2}{c}{\textbf{\emph{mini}ImageNet$\rightarrow$CUB}}\tabularnewline
 \textbf{method} & \textbf{backbone} & \textbf{1-shot} & \textbf{5-shot} & \textbf{1-shot} & \textbf{5-shot}\tabularnewline
\hline

 UBC-FSL (Ours) & ResNet-12 & 41.3$\pm$0.5 & 59.1$\pm$0.6 & 60.2$\pm$0.7 & 80.1$\pm$0.4\tabularnewline

 UBC-FSL (Ours) & ResNet-50 & 41.4$\pm$0.6 & 58.5$\pm$0.6 & 58.8$\pm$0.6 & 79.0$\pm$0.5\tabularnewline

 UBC-FSL (Ours) & ResNet-101 & 42.3$\pm$0.5 & 59.9$\pm$0.6 & 60.8$\pm$0.6 & 80.7$\pm$0.4\tabularnewline

 FSL baseline & ResNet-12 & 46.0$\pm$0.6 & 63.7$\pm$0.5 & 62.4$\pm$0.6 & 79.1$\pm$0.4\tabularnewline

 FSL baseline & ResNet-50 & 46.3$\pm$0.6 & 64.9$\pm$0.5 & 63.2$\pm$0.8 & 79.9$\pm$0.5\tabularnewline

 FSL baseline & ResNet-101 & \second{47.3$\pm$0.6} & 65.6$\pm$0.5 & 64.6$\pm$0.7 & 81.1$\pm$0.5\tabularnewline

 Combined (Ours) & ResNet-12 & 46.4$\pm$0.6 & 65.1$\pm$0.5 & 65.5$\pm$0.6 & 83.0$\pm$0.4\tabularnewline

 Combined (Ours) & ResNet-50 & 47.0$\pm$0.4 & \second{66.3$\pm$0.5} & \second{65.7$\pm$0.8} & 83.2$\pm$0.4\tabularnewline

 Combined (Ours) & ResNet-101 & \first{47.6$\pm$0.6} & \first{67.3$\pm$0.5} & \first{67.4$\pm$0.5} & \first{84.5$\pm$0.4}\tabularnewline
 
 UBC-TFSL (Ours) & ResNet-12 & 41.5$\pm$0.5 & 59.2$\pm$0.6 & 61.1$\pm$0.6 & 81.1$\pm$0.5\tabularnewline

 UBC-TFSL (Ours) & ResNet-50 & 43.1$\pm$0.5 & 61.0$\pm$0.7 & 62.3$\pm$0.6 & 82.8$\pm$0.4\tabularnewline

 UBC-TFSL (Ours) & ResNet-101 & 44.0$\pm$0.6 & 61.7$\pm$0.6 & 63.0$\pm$0.6 & \second{83.3$\pm$0.4}\tabularnewline
\hline 
\end{tabular}
\caption{\textbf{Top-1 accuracies(\%) for cross-domain FSL.} We report the mean of 1000 randomly generated test episodes as well as the 95\% confidence intervals. The top results are highlighted in \first{blue} and the second-best results in \second{green}. Note that for here UBC-TFSL only have additional access to the unlabeled images of the novel classes on source domain.}
\label{tab:transfer1}
\end{table*}

\begin{table*}[htbp]
    \centering
    \small
    
\begin{tabular}{clcccc}
\hline 
  & &  \multicolumn{2}{c}{
\textbf{\emph{tiered}ImageNet$\rightarrow$Caltech}
} & \multicolumn{2}{c}{\textbf{\emph{tiered}ImageNet$\rightarrow$CUB}}\tabularnewline
 \textbf{method} & \textbf{backbone} & \textbf{1-shot} & \textbf{5-shot} & \textbf{1-shot} & \textbf{5-shot}\tabularnewline
\hline

 UBC-FSL (Ours) & ResNet-12 & 46.3$\pm$0.5 & 64.3$\pm$0.6 & 65.2$\pm$0.7 & 84.1$\pm$0.5\tabularnewline

 UBC-FSL (Ours) & ResNet-50 & 58.1$\pm$0.7 & 76.3$\pm$0.6 & 76.9$\pm$0.5 & 91.3$\pm$0.4\tabularnewline

 UBC-FSL (Ours) & ResNet-101 & 57.0$\pm$0.7 & 75.4$\pm$0.6 & 78.9$\pm$0.8 & 92.5$\pm$0.4\tabularnewline

 FSL baseline & ResNet-12 & 63.6$\pm$0.7 & 82.4$\pm$0.5 & 75.2$\pm$0.7 & 90.0$\pm$0.4\tabularnewline

 FSL baseline & ResNet-50 & 70.0$\pm$0.5 & 85.5$\pm$0.5 & 79.5$\pm$0.7 & 91.9$\pm$0.5\tabularnewline

 FSL baseline & ResNet-101 & \first{72.9$\pm$0.7} & \second{87.2$\pm$0.5} & 80.7$\pm$0.7 & 92.6$\pm$0.3\tabularnewline
 
 Combined (Ours) & ResNet-12 & 58.8 $\pm$0.8 & 79.2$\pm$0.6 & 75.6$\pm$0.6 & 90.6$\pm$0.3\tabularnewline

 Combined (Ours) & ResNet-50 & 69.0$\pm$0.7 & 86.2$\pm$0.4 & \second{82.3$\pm$0.6} & \second{93.8$\pm$0.4}\tabularnewline

 Combined (Ours) & ResNet-101 & \second{70.6$\pm$0.7} & \first{87.3$\pm$0.3} & \first{83.8$\pm$0.5} & \first{94.6$\pm$0.3}\tabularnewline
 
 UBC-TFSL (Ours) & ResNet-12 & 44.8$\pm$0.6 & 62.7$\pm$0.7 & 66.0$\pm$0.7 & 84.6$\pm$0.7\tabularnewline

 UBC-TFSL (Ours) & ResNet-50 & 56.5$\pm$0.6 & 74.5$\pm$0.6 & 78.3$\pm$0.6 & 91.9$\pm$0.4\tabularnewline

 UBC-TFSL (Ours) & ResNet-101 & 59.4$\pm$0.7 & 76.3$\pm$0.7 & 81.1$\pm$0.7 & 93.3$\pm$0.5\tabularnewline

\hline 
\end{tabular}
\caption{\textbf{Top-1 accuracies(\%) for cross-domain FSL.} We report the mean of 1000 randomly generated test episodes as well as the 95\% confidence intervals. The top results are highlighted in \first{blue} and the second-best results in \second{green}. Note that for here UBC-TFSL only have additional access to the unlabeled images of the novel classes on source domain. }
\label{tab:transfer2}
\end{table*}

We report our results for cross-domain FSL in Table \ref{tab:transfer1} and Table \ref{tab:transfer2}. In Table \ref{tab:transfer1}, we show results of learning models on \emph{mini}ImageNet and evaluating them on Caltech-256 and CUB.
In Table \ref{tab:transfer2}, we show results of learning models on \emph{tiered}ImageNet and evaluating them on Caltech-256 and CUB.

\end{document}

%% file: math_commands.tex
\usepackage{amsmath,amsfonts,bm}

\def\eqref#1{equation~\ref{#1}}

\def\1{\bm{1}}

\DeclareMathAlphabet{\mathsfit}{\encodingdefault}{\sfdefault}{m}{sl}
\SetMathAlphabet{\mathsfit}{bold}{\encodingdefault}{\sfdefault}{bx}{n}

